\documentclass[10pt,twocolumn,letterpaper]{article}
\usepackage{duckuments}

\usepackage[pagenumbers]{cvpr} 

\definecolor{cvprblue}{rgb}{0.21,0.49,0.74}
\usepackage[pagebackref,breaklinks,colorlinks,allcolors=cvprblue]{hyperref}

\def\paperID{9008} 
\def\confName{CVPR}
\def\confYear{2025}

\title{Graph Canvas for Controllable 3D Scene Generation}

\makeatletter
\def\thanks#1{\protected@xdef\@thanks{\@thanks
        \protect\footnotetext{#1}}}
\makeatother

\author{Libin Liu\textsuperscript{\rm 1, 7}, Shen Chen\textsuperscript{\rm 2}, Sen Jia\textsuperscript{\rm 3}, Jingzhe Shi\textsuperscript{\rm 4}, Zhongyu Jiang\textsuperscript{\rm 7}, \\ \ Can Jin\textsuperscript{\rm 5}, Zongkai Wu\textsuperscript{\rm 6}, Jenq-Neng Hwang\textsuperscript{\rm 7}, Lei Li\textsuperscript{\rm 7,8 \dag} \thanks{\(^{\dag}\) Corresponding Author. (\href{mailto:lilei@di.ku.dk}{\color{black}{\texttt{lilei@di.ku.dk}}}) \\
\textsuperscript{\rm 1}Beijing University of Technology \textsuperscript{\rm 2}East China University of Science
\textsuperscript{\rm 3}Shandong University \textsuperscript{\rm 4}Tsinghua University \textsuperscript{\rm 5}Rutgers University \textsuperscript{\rm 6}Fancy Tech
 \textsuperscript{\rm 7}University of Washington  \textsuperscript{\rm 8}University of Copenhagen}
}

\begin{document}
\maketitle
\begin{abstract}
Spatial intelligence is foundational to AI systems that interact with the physical world, particularly in 3D scene generation and spatial comprehension. Current methodologies for 3D scene generation often rely heavily on predefined datasets, and struggle to adapt dynamically to changing spatial relationships. In this paper, we introduce \textbf{GraphCanvas3D}, a programmable, extensible, and adaptable framework for controllable 3D scene generation. Leveraging in-context learning, GraphCanvas3D enables dynamic adaptability without the need for retraining, supporting flexible and customizable scene creation. Our framework employs hierarchical, graph-driven scene descriptions, representing spatial elements as graph nodes and establishing coherent relationships among objects in 3D environments. Unlike conventional approaches, which are constrained in adaptability and often require predefined input masks or retraining for modifications, GraphCanvas3D allows for seamless object manipulation and scene adjustments on the fly. Additionally, GraphCanvas3D supports 4D scene generation, incorporating temporal dynamics to model changes over time. Experimental results and user studies demonstrate that GraphCanvas3D enhances usability, flexibility, and adaptability for scene generation.
\end{abstract}
    
\section{Introduction} \label{sec:introduction}

Spatial intelligence, defined as an AI system’s ability to comprehend, interpret, and manipulate spatial relationships within a given environment, is fundamental to the development of systems capable of effective interaction with physical spaces. Despite considerable advancements in this domain~\cite{janowicz2020geoai, AI-LLM-review, spatial-computing, feng2024layoutgpt, cai2024spatialbot, paschalidou2021atiss}, current methods for 3D layout generation exhibit significant limitations in terms of flexibility, usability, and adaptability, which restrict their utility in dynamic, real-time applications. Many existing approaches ~\cite{mvdream, DreamFusion, GaussianDiffusion} rely on resource-intensive retraining, stringent input configurations, or manually defined masks and constraints, each of which introduces rigidity into the 3D scene generation process, ultimately impacting efficiency and adaptability.

\begin{figure*}[ht]
   \centering
   \includegraphics[width=0.98\linewidth]{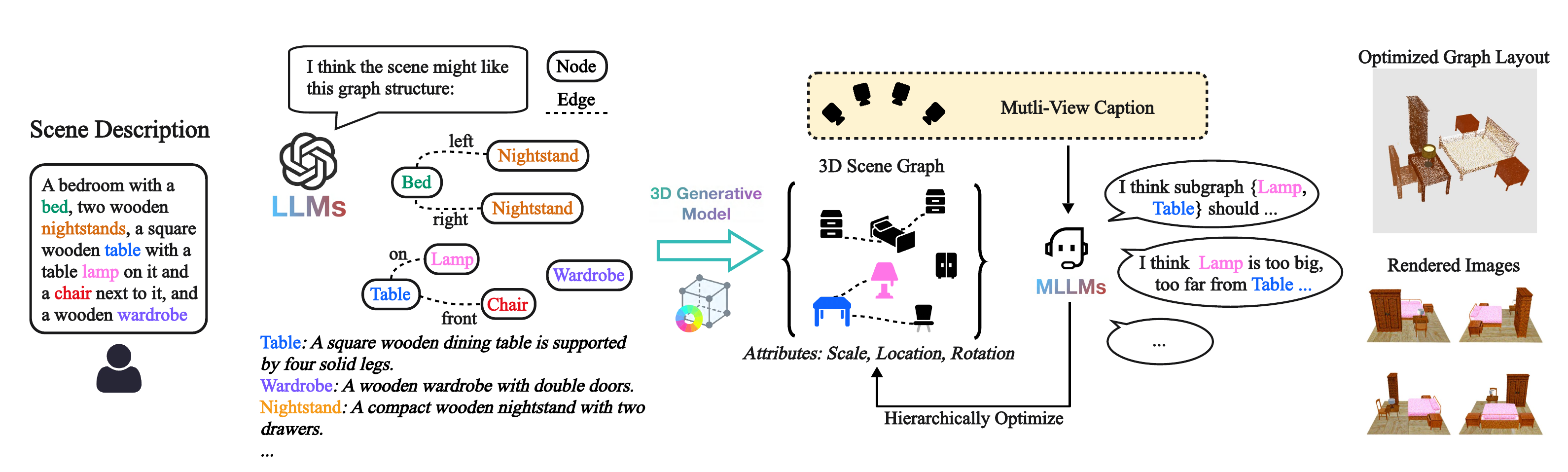}
   \caption{\textbf{Overview of our method. }Given a brief scene description, our method first allows the LLMs to construct a graph structure to manage the objects mentioned in the scene prompt and the relationships between them. Additionally, each object in the scene is provided with a richer description and passes through a 3D generative model to create corresponding 3D objects. We capture views of these 3D objects and let the MLLMs analyze whether the relative positions between objects are accurate. Ultimately, we achieve excellent results in terms of scene layout and rendering quality. }
   \label{fig:pipeline}
 \end{figure*}

Existing frameworks, such as LayoutGPT~\cite{feng2024layoutgpt} and similar layout generation models ~\cite{gupta2021layouttransformer, chen2024sculpt3d, zhou2024layout}, exemplify these constraints. While effective in generating initial 3D layouts, these models often require extensive manual intervention or detailed scene specifications whenever environmental changes are introduced. This requirement renders them impractical for applications demanding continuous or real-time adaptability. Moreover, their inherent inflexibility often necessitates frequent retraining to accommodate novel scenarios, a process that is both computationally expensive and time-consuming. Consequently, these models exhibit limited transferability and generalizability across diverse environmental contexts.

To address these challenges, we introduce \textbf{GraphCanvas3D}, a novel framework that aims to bridge the limitations of existing 3D layout generation methodologies by offering a programmable, extensible and transferable paradigm for 3D scene construction. GraphCanvas3D adopts a modular, Graph-based approach, representing spatial relationships through graph structures and utilizing Graph to link real-world entities within a cohesive 3D representation. This approach provides users with a flexible programming interface to create, modify, and expand 3D scenes across various environments, effectively eliminating the need for specialized retraining or intricate scene definitions.

Another important capability of GraphCanvas3D is its support for dynamic scene generation without requiring retraining or manual reconfiguration. This flexibility enables real-time, interactive scene editing, allowing users to modify scenes using concise natural language instructions. By leveraging a graph-based structure that adapts based on contextual text inputs, GraphCanvas3D supports precise, responsive adjustments to 3D layouts, allowing seamless component modifications. 
GraphCanvas3D incorporates time-based adjustments within its graph structure, enabling the creation of temporally evolving 3D scenes. This design allows objects and their spatial relationships to continuously evolve, supporting coherent 4D environments with minimal user intervention. Traditional methods often require extensive retraining, preconfigured datasets, or manual setup for each modification, making real-time 4D scene adaptation infeasible. With Multimodal Large Language Models (MMLM) driven, GraphCanvas3D’s graph-based pipeline maintains temporal coherence and adaptability, establishing a new standard for flexible, real-time 3D and 4D scene generation. The contributions of this work are as follows: 

\begin{enumerate}
    \item We introduce a hierarchical, Graph-based, off-the-shelf framework for 3D scene generation that is both programmable and extensible, eliminating the need for retraining or manually specified scene details.
    \item Our generated graph is flexible and editable, supporting adaptive, real-time scene generation and enabling dynamic modification of 3D layouts.
    \item Extensive experimental evaluations and user studies demonstrate that GraphCanvas3D outperforms state-of-the-art methods in terms of usability, flexibility, and adaptability across diverse application scenarios.
\end{enumerate}

\section{Related Work}
\label{sec:related}

\subsection{3D Representations}
3D representations can be broadly categorized into implicit and explicit forms. Neural Radiance Fields (NeRF) \cite{Nerf2021} exemplify a widely used implicit approach, mapping 3D coordinates and viewing directions to color and density values via a multilayer perceptron (MLP). Mip-NeRF \cite{mipnerf2021} enhances NeRF by using anti-aliased conical frustums, which effectively address aliasing and improve detail handling, leading to higher-quality rendered images. SparseNeRF \cite{sparsenerf2023} builds on this by incorporating depth information to reduce reliance on densely sampled input images, enabling high-quality 3D reconstruction from sparse views. However, NeRF-based methods typically demand substantial computational resources, limiting their scalability and applicability in real-time scenarios.

In contrast, 3D Gaussian Splatting (3DGS) \cite{3dgs} offers an efficient, explicit 3D representation by optimizing Gaussian spheres to capture the 3D environment. Scaffold-GS \cite{scaffold2024} enhances 3DGS by employing anchor points for the efficient distribution of local Gaussians, adjusting properties based on viewing direction and distance. Mip-Splatting \cite{mipsplatting2024} further improves reconstruction quality by controlling the frequency of 3D Gaussians, thus enhancing detail retention and enabling more efficient rendering.

\subsection{Text-to-3D Generation}
NeRF-based methods have played a crucial role in advancing text-to-3D generation, transforming textual prompts into 3D representations. DreamFusion \cite{DreamFusion} and Magic3D \cite{Magic3D} employ diffusion models to optimize NeRF for single-object synthesis. Despite their success in generating standalone objects, these techniques encounter limitations when scaling to multi-object scenes. ProlificDreamer \cite{ProlificDreamer} improves 3D fidelity by integrating shape priors but struggles with inter-object interactions. Comp3d \cite{Comp3d} and CompoNeRF \cite{CompoNeRF} approach multi-object scenes with layout-constrained NeRF, though they require manual setup and may lead to visual artifacts. Recently, methods integrating 3DGS with diffusion models have been proposed to accelerate text-to-3D generation. For example, approaches by Yi et al. \cite{text-to-3dgs} and Liang et al. \cite{Diffusion-text-to-point} use text-to-point models to initialize 3DGS with human priors, while others \cite{GaussianCube,dreamgs,Two-stage-3DGS,gs-two-stage-3d-generation} adopt two-stage optimization for geometry and texture. Although 3DGS offers speed advantages, multi-object scenes still present challenges due to weak layout constraints, leading to geometric inconsistencies and visual drift in scene content.

Recently, Large Language Models (LLMs) \cite{3d-edit-llm,avator-llm,procedural-3D-LLM} have been explored for their capacity in spatial reasoning, assisting 3D generation by interpreting text prompts to discern object relationships and support spatial layouts. LayoutGPT \cite{feng2024layoutgpt} advances this field by providing a CSS-like syntax for detailed layout control, improving spatial configuration specificity. SceneWiz3D \cite{SceneWiz3D} combines LLMs with layout-based NeRF to optimize scene composition, while GALA3D \cite{gala3d} uses LLMs for initial layout creation, employing layout-guided 3D Gaussian representation with adaptive constraints to refine geometry and inter-object interactions, thus achieving coherent multi-object 3D scenes. Nonetheless, LLM-based approaches often face challenges with spatial ambiguity, resulting in misaligned or floating objects due to imprecise layout generation. To address these issues, our method incorporates adaptive layout-guided Gaussian modeling, refining LLM-initialized layouts to improve spatial coherence and deliver consistent, high-quality 3D representations for complex, multi-object scenes.

\section{Method}

\begin{figure*}[ht]
   \centering
   \includegraphics[width=\linewidth]{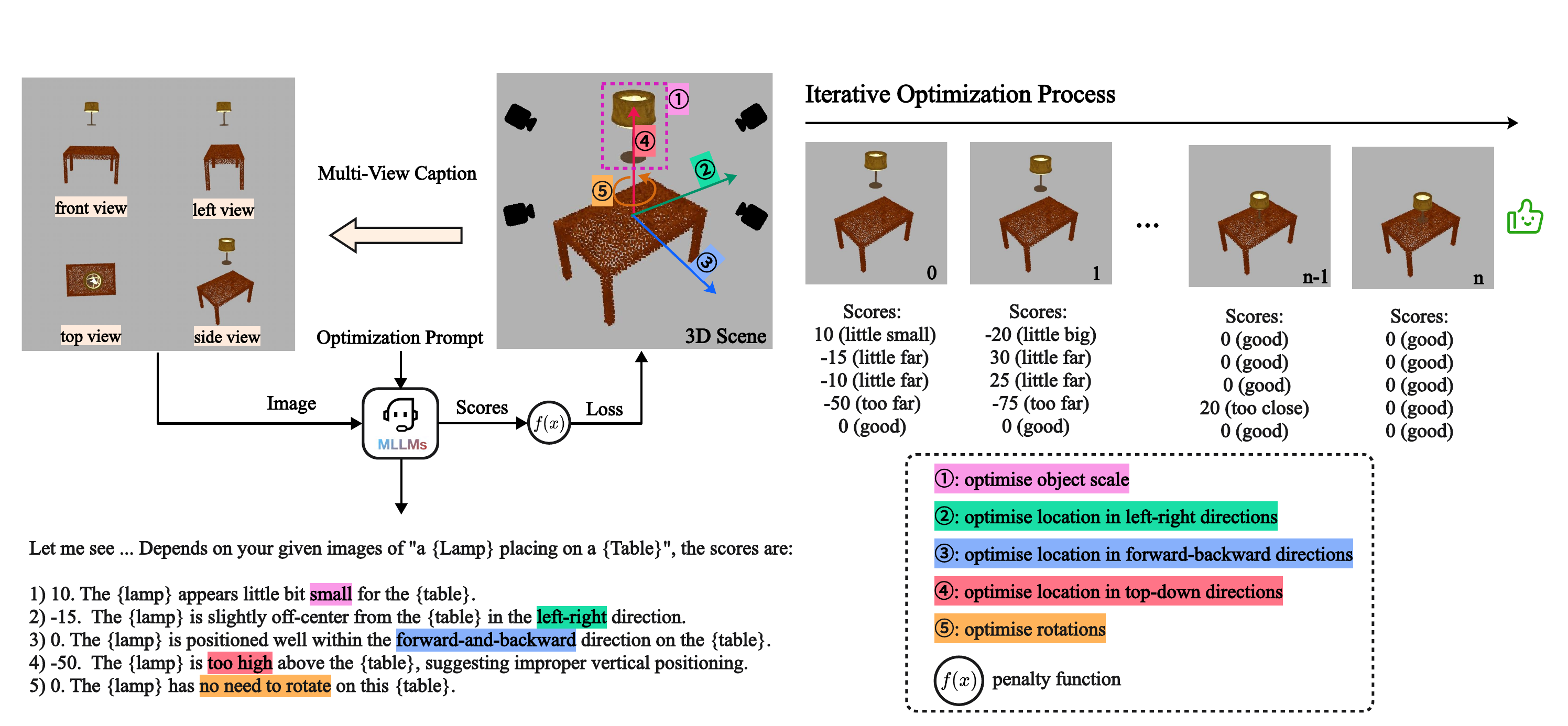}
   \caption{\textbf{Edge Optimization Process. }When optimizing an edge, we capture the 3D scene from four different viewpoints to obtain images from these perspectives. These four images are then sent along with an optimized prompt into the MLLMs, which analyzes the inherent relationships between objects across the images and provides corresponding scores. These scores serve as references for optimizing this edge. After passing through penalty function, the scores are propagated to the scene, guiding iterative optimization. }
   \label{fig:edge-optim}
 \end{figure*}
 
\label{sec:Method}

\subsection{Problem Formulation}

In GraphCanvas3D, given a scene prompt, we employ a large language model (LLM) to parse the input, identify objects, and infer both explicit and implicit spatial relationships. Each identified object \( o_i \) is represented by a feature vector \( \mathbf{f}_i \) with components encoding its spatial and geometric attributes:

\[
\mathbf{f}_i = [x_i, y_i, z_i, s_i, r_i],
\]

where:
\begin{itemize}
    \item \( (x_i, y_i, z_i) \) represents the object's 3D spatial position,
    \item \( s_i \) denotes the scale factor, and
    \item \( r_i \) is the rotation factor along the z-axis, an essential attribute for orientational consistency and scene coherence.
\end{itemize}

We frame the 3D scene generation task as an optimization problem over a structured graph \( \mathcal{G} \), where:
\begin{enumerate}
    \item \textbf{Nodes} represent individual objects in the scene, each characterized by a feature vector \( \mathbf{f}_i \) that captures its 3D properties,
    \item \textbf{Edges} denote spatial relationships between objects, derived from linguistic and contextual cues in the scene prompt.
\end{enumerate}

Our objective is to determine an optimal graph configuration \( \mathcal{G}^* \) that reconstructs the 3D scene in alignment with human spatial intuition. This is achieved by minimizing a global energy function \( E_{\text{scene}} \), defined as follows:

\begin{equation}
\mathcal{G}^* = \arg \min_{\mathcal{G}} E_{\text{scene}},
\end{equation}

where \( E_{\text{scene}} \) quantifies deviations from the desired spatial configurations and relationships as inferred from the input prompt.

\subsection{Overview}
As shown in Fig~\ref{fig:pipeline}, We propose a hierarchical, programmable paradigm for 3D layout generation that circumvents the need for predefined object specifications, such as external files containing object dimensions or positions within the 3D scene. Instead, our approach enGraphs object attributes and their interrelationships through a series of parameterized functions. This paradigm facilitates the generation of coherent 3D scene layouts from succinct textual descriptions, followed by high-quality rendering. Furthermore, our layout paradigm supports iterative modifications of previously rendered scenes, enabling users to add new objects that automatically integrate into the existing layout, or to seamlessly remove or reposition elements without disrupting scene coherence.

The core of our methodology centers on a programmable graph that orchestrates the scene layout. Individual objects are represented as nodes, denoted by $o_i$, and spatial relationships between objects are defined as edges, represented by $l_{ij} = \zeta(o_i, o_j)$, where $\zeta$ is an edge-level optimization function that enGraphs the relationship between objects $o_i$ and $o_j$. We formalize the graph as $\mathcal{G} = (\mathcal{V}, \mathcal{E})$, where $\mathcal{V} = \{ o_1, o_2, \dots, o_N \}$ represents the set of nodes, and $\mathcal{E} = \{ l_1, l_2, \dots, l_M \}$ denotes the set of edges connecting these nodes. 

To enhance robustness, we first construct a collection of subgraphs, each consisting of a set of connected nodes, denoted by $G_i = F_s(\mathcal{V}_i, \mathcal{E}_i)$, where $F_s$ is the subgraph-level optimization function that ensures local consistency. These subgraphs are subsequently aggregated to form the complete graph $\mathcal{G} = F_g(\{G_i\})$, with $F_g$ representing the global optimization function responsible for ensuring overall coherence.

\subsection{Edge-Level Optimization}

To ensure spatial coherence among connected objects, GraphCanvas3D employs an iterative edge-level optimization strategy, refining the spatial relationships between pairs of connected objects \( (o_i, o_j) \) in the scene. As illustrated in Figure~\ref{fig:edge-optim}, this optimization process minimizes deviations from ideal configurations by aligning relationships with high-level semantic expectations derived from the scene prompt. Each pair is evaluated based on an edge cost function \( \zeta(o_i, o_j) \), which takes into account relative positions, scales, and orientations.

The structured edge-level optimization process is outlined as follows:

\begin{enumerate}
    \item \textbf{Multi-View Rendering:} For each pair of connected objects, we generate four distinct views of the subgraph containing the target objects, capturing the scene from the front, left, top, and an oblique perspective. These multi-view representations provide the multimodal LLM with a comprehensive set of visual cues, allowing for accurate assessment of spatial relationships.

    \item \textbf{Scoring via LLM Query:} A predefined edge prompt is used to query the LLM, which assesses the appropriateness of the relative positions, scales, and orientations of the objects across the generated views. The LLM outputs a set of scores \( \mathbf{s}_{ij} = [s_{ij}^1, s_{ij}^2, s_{ij}^3, s_{ij}^4, s_{ij}^5] \), where each score \( s_{ij}^k \) (for \( k=1, \dots, 5 \)) quantifies the spatial adequacy of the objects’ arrangement in each view on a scale from \([-100, 100]\).

    \item \textbf{Loss Computation:} These scores are transformed into a loss value \( L_{ij} \) that guides the optimization of the edge. Specifically, the scores \( \mathbf{s}_{ij} \) are passed through activation functions tailored for scale, translation, and rotation adjustments, yielding directional loss gradients. The total loss for the edge \( l_{ij} \) is computed as a weighted sum:
    \begin{equation}
    L_{ij} = \sum_{k=1}^5 w_k \cdot f(s_{ij}^k),
    \end{equation}
    where \( w_k \) represents the weight assigned to each view, and \( f(s_{ij}^k) \) is a penalty function that increases the loss for deviations from the target spatial relationships.

    \item \textbf{Gradient-Based Updates:} Using the computed loss \( L_{ij} \), a gradient descent update is applied to adjust the feature vectors of the connected nodes. This update rule is expressed as:
    \begin{equation}
    \mathbf{f}_i \leftarrow \mathbf{f}_i - \eta \frac{\partial L_{ij}}{\partial \mathbf{f}_i},
    \end{equation}
    where \( \eta \) is the learning rate. To maximize computational efficiency, only the incoming vertices are directly optimized. This selective adjustment allows indirect propagation of spatial coherence through adjacent vertices, without re-evaluating the entire graph structure.

    \item \textbf{Convergence Check:} The optimization loop continues iteratively until the loss \( L_{ij} \) reaches a predefined threshold, signifying that the spatial relationship has achieved the required coherence.

\end{enumerate}


\subsection{Subgraph-Level Optimization}

Following the completion of edge-level optimizations, which establish robust spatial relationships between connected objects, the next stage involves assembling these optimized objects into coherent subgraphs and, subsequently, a unified global scene. Each subgraph \( G_i \) is optimized independently to ensure internal spatial coherence, laying the foundation for an integrated scene that aligns with high-level semantic configurations.

\textbf{Independent Subgraph Optimization:} Each subgraph is constructed by grouping objects with strong inter-object relationships, typically defined by closely aligned spatial attributes and semantic associations. These subgraphs \( G_i \) are optimized to minimize internal energy functions \( E_{subgraph}(G_i) \), ensuring that objects within each subgraph maintain coherent relative positions, scales, and orientations. This step ensures that local regions of the scene exhibit spatial fidelity and that subgraphs can be integrated without internal inconsistencies.

\textbf{LLM-Guided Subgraph Placement:} Once subgraphs achieve local optimization, their placements within the overall scene are guided by the large language model (LLM), which interprets high-level prompts that specify details about the subgraphs, such as the number of constituent objects, their inferred sizes, and interrelationships. The LLM uses this contextual information to propose initial placements that reflect semantic and spatial expectations at the scene level. This guided placement ensures that the relationships between different subgraphs are consistent with the scene’s intended spatial semantics.

\subsection{Graph-Level Optimization}
To achieve a globally coherent scene layout, a higher-level optimization function \( F_g \) is employed. This function refines the spatial arrangements of subgraphs, minimizing the global objective:

\begin{equation}
\mathcal{G}^* = \arg \min_{\mathcal{G}} \sum_{i=1}^K E_{subgraph}(G_i) + \sum_{(G_p, G_q) \in \mathcal{E}_g} \psi(G_p, G_q),
\end{equation}

where \( \psi(G_p, G_q) \) represents the penalty function applied to any misalignment or inappropriate spacing between adjacent subgraphs \( G_p \) and \( G_q \). This term enforces spatial consistency between subgraphs, preserving both the relative positioning and semantic alignment across the global scene.


\begin{figure*}[ht]
   \centering
   \includegraphics[width=\linewidth]{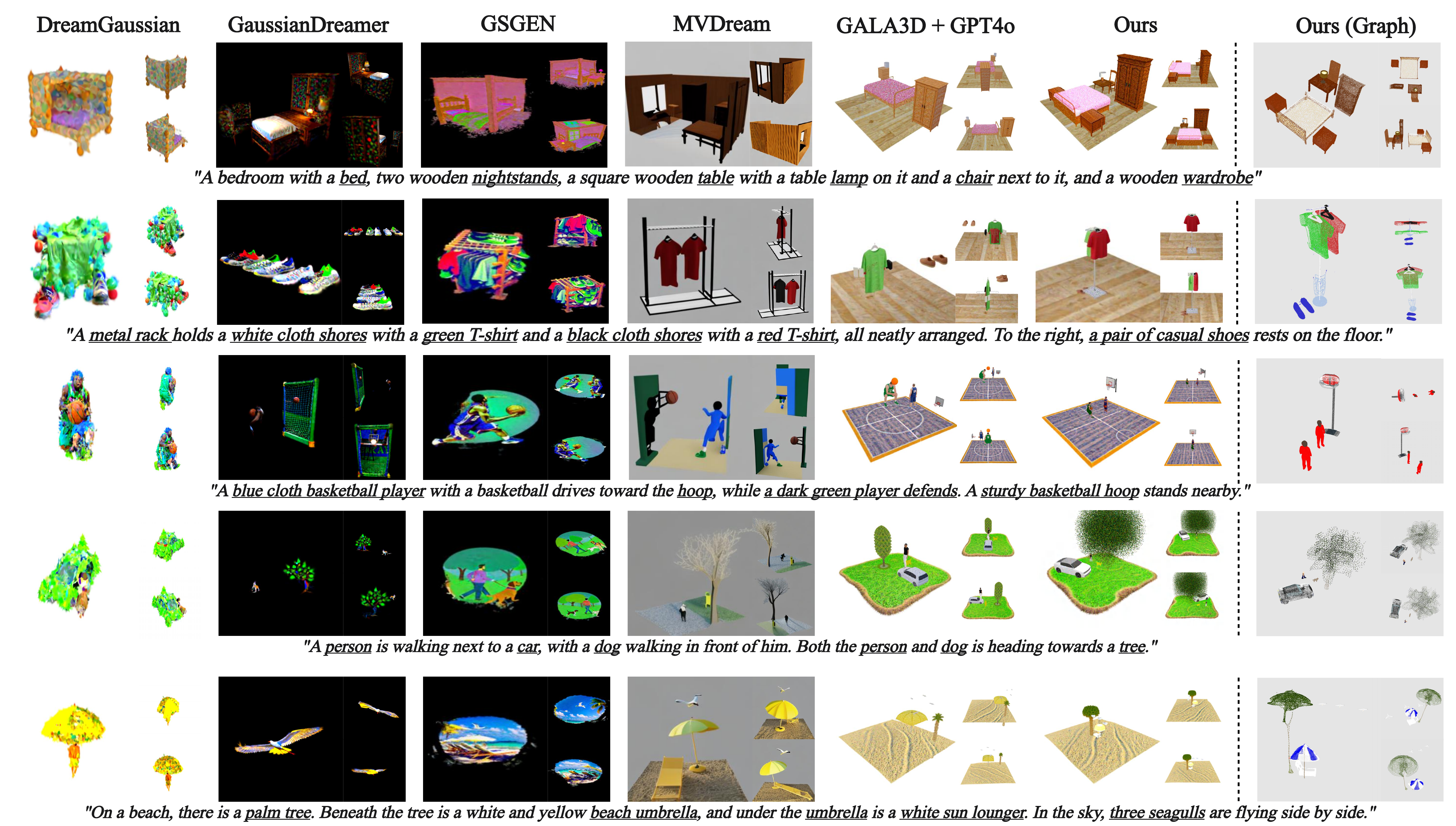}
   \caption{\textbf{Qualitative Comparisons of Text-to-3D Scene Generation Approaches.} Our method generates high-quality, interactive multi-object scenes and complex compositions that closely follow input textual descriptions. In the final column of the figure, we present the graph structure of the GraphCanvas3D method before rendering. GraphCanvas3D’s graph structure represents the 3D scene with nodes for objects and edges for their spatial relationships, ensuring consistency and coherence in scene generation.}
   \label{fig:experiment}
 \end{figure*}
 
\subsection{Dynamic Scene Modification}



GraphCanvas3D supports dynamic scene modifications (4D scene), allowing for moving, adding, removing, and repositioning of objects. For additions, a new node \( o_{\text{new}} \) is introduced with spatial relationships established through new edges \( l_{\text{new}, i} \), which are optimized for coherence. In the case of removals, the corresponding node and its edges are deleted, followed by re-optimization of adjacent nodes to maintain alignment. Repositioning involves updating the feature vector \( \mathbf{f}_i \) of the target node and re-optimizing related edges and subgraphs to preserve the overall scene layout.

After the initial layout is constructed, high-quality rendering enhances the visual fidelity of scenes. GraphCanvas3D use diffusion-based methods to provide high-resolution and stylistic consistency rendering result for every object and the entire scene. This multi-step rendering pipeline allows GraphCanvas3D to produce realistic, unified visual representations, supporting applications in novel view synthesis and interactive environments.

\section{Experimental Results}

\begin{table}[t]
    \centering
    \begin{tabular}{lcc}
        \toprule
        \textbf{Methods} & \textbf{CLIP Score} & \textbf{MLLM Score}\\
        \midrule
        DreamGaussian    & 22.33& 1.7\\
        GaussianDreamer  & 26.17& 3.0\\
        MVDream          & 26.25& 4.4\\
        GS-Gen           & 26.28& 4.1\\
        GALA3D           & 28.67& 7.0\\
        GraphCanvas3D& \textbf{29.67}& \textbf{8.3}\\
        \bottomrule
    \end{tabular}
    \caption{\textbf{Comparison of Methods by CLIP Score and MLLM Sort Score.} We evaluate our rendering results using CLIP and a multimodal language model, and observe that, our approach outperforms previous methods.}
    \label{tab:clip+mllm}
\end{table} 
\textbf{Implementaion details.} In our experiments, we utilized ChatGPT-4o~\cite{chatgpt} as both the Large Language Model (LLM) and the Multimodal Language Model (MLLM), alongside Point-E~\cite{pointe} as the 3D generative model. The Point-E model generates 4096-point clouds, providing a foundational approximation of object contours, though it lacks high-resolution detail. To enhance the fidelity of these representations, we expand each object’s point cloud to 100,000 points via bilinear interpolation. These enriched point clouds are then employed as the initialization for each object’s 3D Gaussian Splatting (3DGS) process. To enhance texture and detail in each object, we utilized MVDream~\cite{mvdream} for object rendering. To maintain consistency across the entire scene, we employed ControlNet~\cite{zhang2023controlnet} for comprehensive scene rendering, ensuring seamless integration of objects within the overall environment. We set the MVDream guidance scale to 7.5 to preserve object structural integrity while enhancing texture details during rendering. In our 3D Gaussian Splatting (3DGS) framework, parameters such as opacity, position, spherical harmonics coefficients, and covariance are consistent with those in GALA3D~\cite{gala3d}. All experiments were conducted on a single A100 GPU, with approximately 24GB of memory usage.


\begin{table}[t]
    \centering
    \captionsetup{skip=4pt} 
    \resizebox{\columnwidth}{!}{%
    \begin{tabular}{lccc}
        \toprule
        \textbf{Methods} & \textbf{Scene Quality} & \textbf{Geometric Fidelity} & \textbf{Layout Realism}\\
        \midrule
        DreamGaussian    & 5.22 & 4.18 & 4.30 \\
        GaussianDreamer  & 6.09 & 5.71 & 5.23 \\
        MVDream          & 7.32 & 7.98 & 7.07\\
        GS-Gen           & 6.90 & 6.65 & 6.92\\
        GALA3D           & 7.28 & 7.34 & 7.59 \\
        GraphCanvas3D    & \textbf{8.01} & \textbf{8.64} & \textbf{9.02}\\
        \bottomrule
    \end{tabular}%
    }
    \caption{\textbf{User study results.} Comparison of Human evaluation results between GraphCanvas3D and other Text to 3D methods. Participants rated each method based on three metrics. The higher the score, the stronger the preference.}
    \label{tab:user-study}
    \vspace{-10pt} 
\end{table}

\begin{figure*}[ht]
   \centering
   \includegraphics[width=\linewidth]{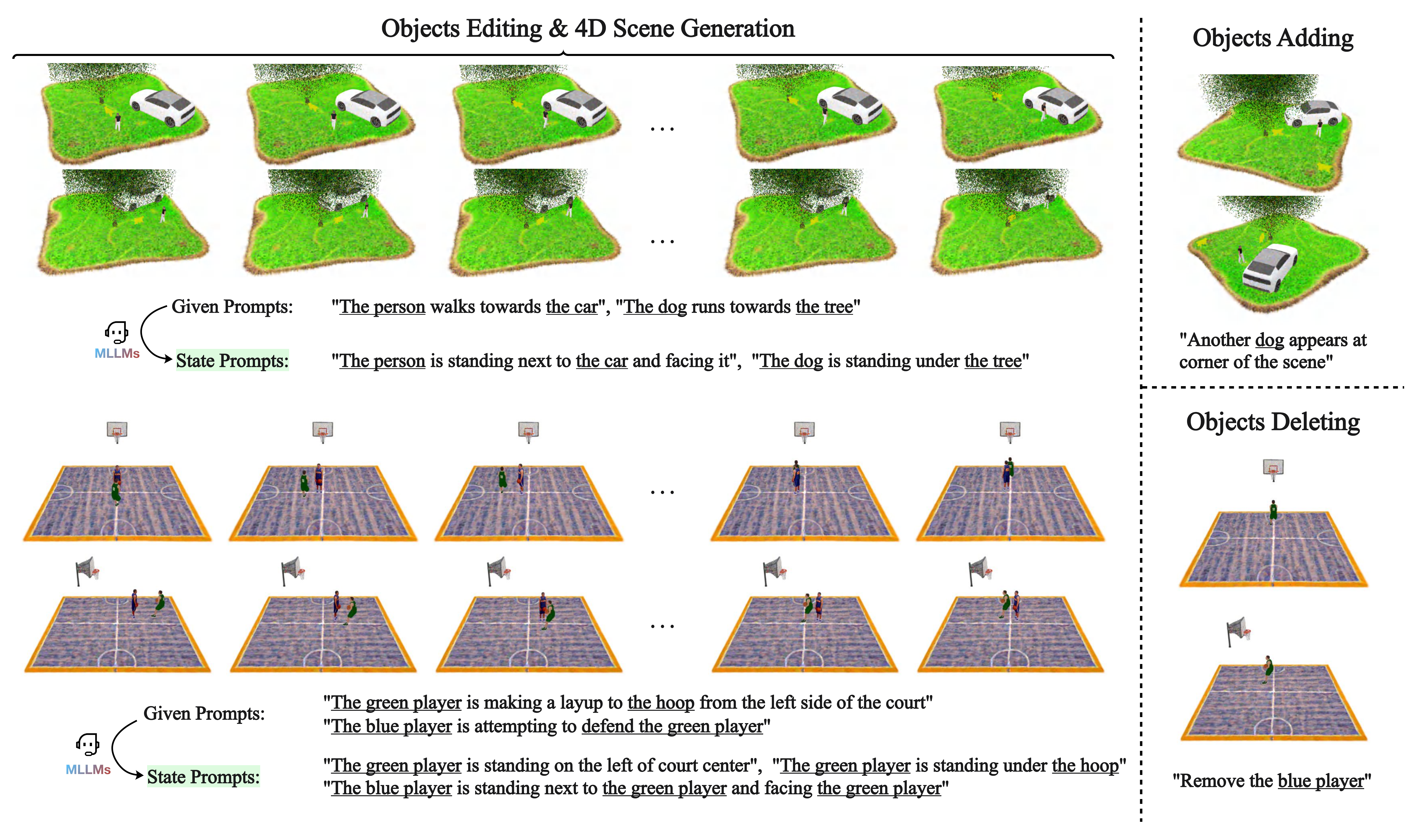}
   \caption{\textbf{Experiments of Dynamic Scene Modification.} GraphCanvas3D is capable of object editing, adding, deleting and 4D scene generation based on textual descriptions. }
   \label{fig:experiment2}
 \end{figure*}
\subsection{Quantitative Comparsion.} To evaluate our approach on the Text-to-3D task, we benchmark against state-of-the-art methods, including DreamGaussian~\cite{dreamgs}, GaussianDreamer~\cite{gsdreamer}, MVDream~\cite{mvdream}, GS-Gen~\cite{gsgen}, and GALA3D~\cite{gala3d}. Following previous studies~\cite{DreamFusion, dreamtime}, we use the CLIP Score to assess the alignment between textual descriptions and generated images. Additionally, we leverage a Multi-modal Large Language Model (MLLM) to evaluate the semantic consistency between scene descriptions and generated images from multiple perspectives. This comprehensive evaluation enables the MLLM to rank and score all methods based on their outputs. As shown in Table \ref{tab:clip+mllm}, our method achieves the highest CLIP and MLLM scores.

\subsection{Qualitative Comparison.} We present a qualitative comparison of Text-to-3D scene generation in Figure \ref{motivation} and Figure \ref{fig:experiment}. Notably, GALA3D requires precise input for each object’s location, scale, and rotation during scene generation. To address this, we adopt a prompt format similar to LayoutGPT~\cite{feng2024layoutgpt}, using ChatGPT-4o to generate these attributes. While we utilized the CSS format from LayoutGPT, our prompts did not include a large number of directly related scene examples. Compared to existing methods, GraphCanvas3D produces scenes with a more realistic and cohesive structure, delivering robust rendering results adaptable to various scenarios. This advantage is attributed to our graph-based framework and optimization-driven scene layout control, which enables our method to outperform others in both quality and adaptability.

\subsection{User Study.} To further assess the effectiveness of our method in generating high-quality, text-consistent 3D scenes, we conducted a user study with 67 participants. The study involved comparing 3D models generated by our approach with those produced by competing methods, using eight distinct text descriptions. Participants evaluated each model across three dimensions: (a) Scene Quality, (b) Geometric Fidelity, and (c) Layout Realism, assigning ratings on a scale from 1 to 10 (with 10 indicating the highest score). As summarized in Table ~\ref{tab:user-study}, our method consistently achieved superior ratings, demonstrating its clear advantage over previous approaches.

\subsection{Dynamic Scene Modification.} As illustrated in Figure ~\ref{fig:experiment2}, our approach facilitates not only the generation of static 3D scenes from text but also supports dynamic editing, addition, and deletion within 3D scenes. Moreover, our method extends to generating 4D scenes that evolve over time. The core logic behind both object editing and 4D scene generation remains consistent in our approach. Given a prompt that describes a transformation process, GraphCanvas3D enables multimodal large language models (MLLMs) to analyze the scene and determine the final state of the objects, referred to as “state prompts.” These state prompts then guide the optimization of our scene graph, where an iterative process yields a temporal transformation sequence, achieving object editing and 4D scene generation in a unified manner. Additionally, objects can be efficiently added or removed within the scene by modifying the scene graph, allowing for flexible and efficient scene adjustments.

\section{Ablation Study}

\begin{figure}[ht]
   \centering
   \includegraphics[width=\linewidth]{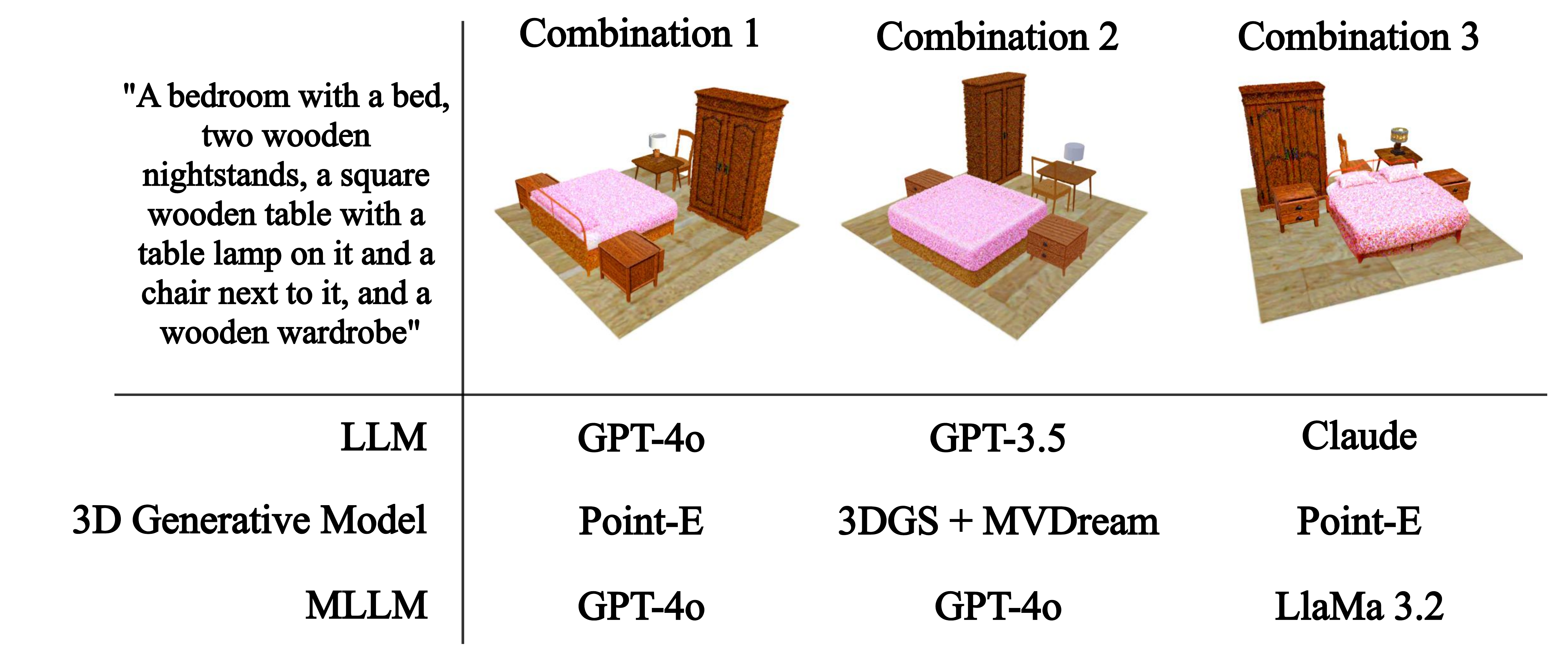}
   \caption{\textbf{Ablation Study of Model Flexibility.} Our method can accomplish text-to-3D scene generation task with various models. We presented three different model selection approaches, all of which achieved promising results.}
   \label{fig:abstudy-diff-models}
 \end{figure}

\textbf{Model Flexibility.} Our method imposes no strict model requirements on the LLM, 3D Generative Model, or MLLM, allowing for flexible integration with various model architectures. To enhance the adaptability of our approach, we examined three different model combinations in this ablation study, as shown in Figure ~\ref{fig:abstudy-diff-models}. At the core of our method is a graph-based structure that encodes objects within a scene and their interrelationships, ensuring consistently reliable outcomes regardless of the specific models employed and underscoring the approach’s robustness and versatility.

\begin{figure}[ht]
   \centering
   \includegraphics[width=0.9 \linewidth]{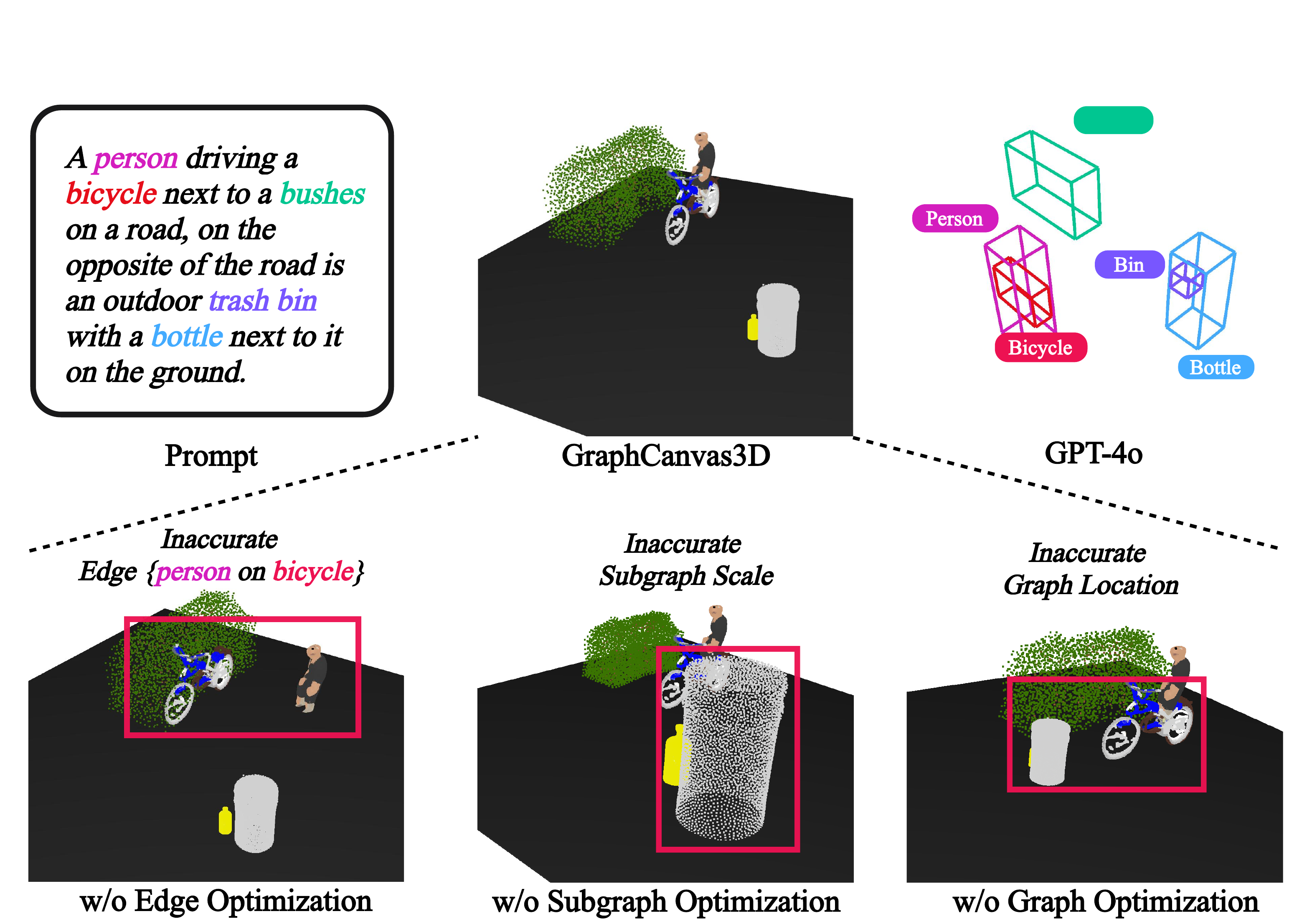}
   \caption{\textbf{Ablation Study of Hierarchical Optimization.} Experiments confirm the effectiveness of each level of our optimization, underscoring its essential role in the GraphCanvas3D framework.}
   \label{fig:abstudy-optim}
 \end{figure}

\textbf{Hierarchical Optimization.} Figure ~\ref{fig:abstudy-optim} presents an ablation study evaluating the effectiveness of our hierarchical optimization approach. Given a prompt, GraphCanvas3D generated a layout that accurately reflects real-world spatial arrangements, whereas the layout produced solely by GPT-4o appeared disorganized. Removing edge optimization resulted in notable misalignment between the person and bicycle, emphasizing the role of edge optimization in maintaining spatial constraints between connected nodes. Without subgraph optimization, substantial scale discrepancies emerged between two subgroups—one containing the person, bicycle, and bushes, and the other containing the trash can and bottle—resulting in an unrealistic layout. Additionally, when graph optimization was excluded, the generated scene lacked overall coherence, with nearly all objects clustered on one side, contradicting the prompt’s intended layout. This study demonstrates the reliability of our hierarchical optimization approach and the essential role of each optimization level in achieving coherent scene generation.

\section{Conclusion}

In this work, we introduced \textbf{GraphCanvas3D}, a novel framework that addresses the limitations of 3D scene generation methods by providing a flexible, modular, and adaptive approach to 3D scene construction. Distinct from prior models, GraphCanvas3D employs Multi-Layered Language Models (MLLMs) to enable real-time scene manipulation through natural language descriptions, obviating the need for retraining or rigid input configurations. By representing spatial relationships as graph structures, GraphCanvas3D offers an intuitive interface that facilitates dynamic scene modifications, significantly enhancing usability and adaptability across diverse environments. Our experimental results validate GraphCanvas3D’s effectiveness, demonstrating superior performance in flexibility, responsiveness, and user-centered design when compared with existing approaches, thus underscoring its potential as a robust tool for applications requiring real-time adjustments and spatial intelligence. Future work will focus on further scaling GraphCanvas3D's capabilities and efficiency, including integration with virtual and augmented reality platforms to extend its utility in interactive and adaptive 3D scene generation.

{
    \small
    \bibliographystyle{ieeenat_fullname}
    \bibliography{main}
}
\clearpage
\setcounter{page}{1}
\maketitlesupplementary

To provide a comprehensive understanding of our method, this supplementary section elaborates on key components such as graph construction, edge optimization, subgraph optimization, and final graph placement. Detailed explanations and additional examples are provided to enhance clarity.

\section{Optimized Processing}

\begin{figure*}[h]
   \centering
   \includegraphics[width=1.0\linewidth ]{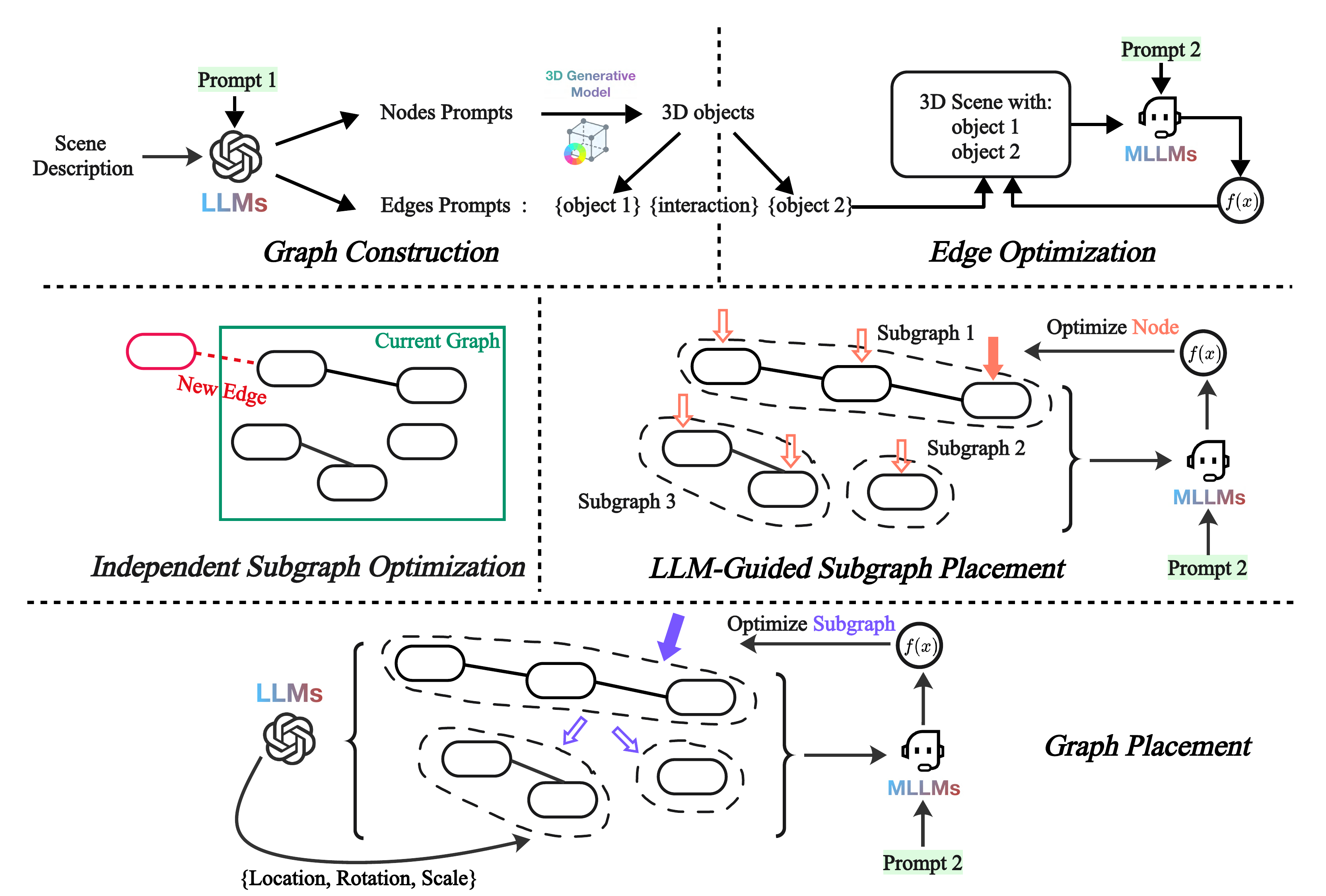}
   \caption{\textbf{Detailed Overview of our Method.} In our appendix, we provide an extended version of the overview diagram of our method, accompanied by a more detailed explanation.}
   \label{fig:appendix-overview}
 \end{figure*}

We provide an expanded explanation of the methodology, as illustrated in the Figure ~\ref{fig:appendix-overview}. In the following paragraphs, each subsection of the methods will be further elaborated in greater detail.

\textbf{Graph Construction.} We designed Prompt 1 (shown in Table ~\ref{tab:template_prompts})  to guide LLMs in performing instance-level segmentation of objects and relationships with a scene description $T_s$, resulting in the generation of node prompts and edge prompts. Each instance object is represented as a vertex in the graph, containing attributes such as its 3D representation (e.g., point cloud), scale, position, and rotation. Each instance relationship is represented as an edge in the graph, defined by two connected vertices and a directed relationship attribute.

\textbf{Edge Optimization.} For each edge inferred by the LLMs, we perform an edge optimization process. Here, we provide additional details about the optimization. During each optimization step, three sets of objects are involved: the entire set of objects  $X_{\text{all}}$ , the object being optimized  $X_1$, and the remaining objects  $X_2 = X_{\text{all}} - X_1$ . The optimization algorithm captures images of these objects from four different viewpoints, which are then input into the MLLMs along with Prompt 2 (shown in Table ~\ref{tab:template_prompts}) to evaluate their scores. During edge optimization,  $X_1$  represents the source node object of the edge (in-degree),  $X_2$  represents the node vertex object of the edge (out-degree), and  $X_{\text{all}}$  represents the combined context of both objects. Our edge optimization is able to establish reasonable relationships edge.

\begin{figure}[h]
   \centering
   \includegraphics[width=0.6 \linewidth]{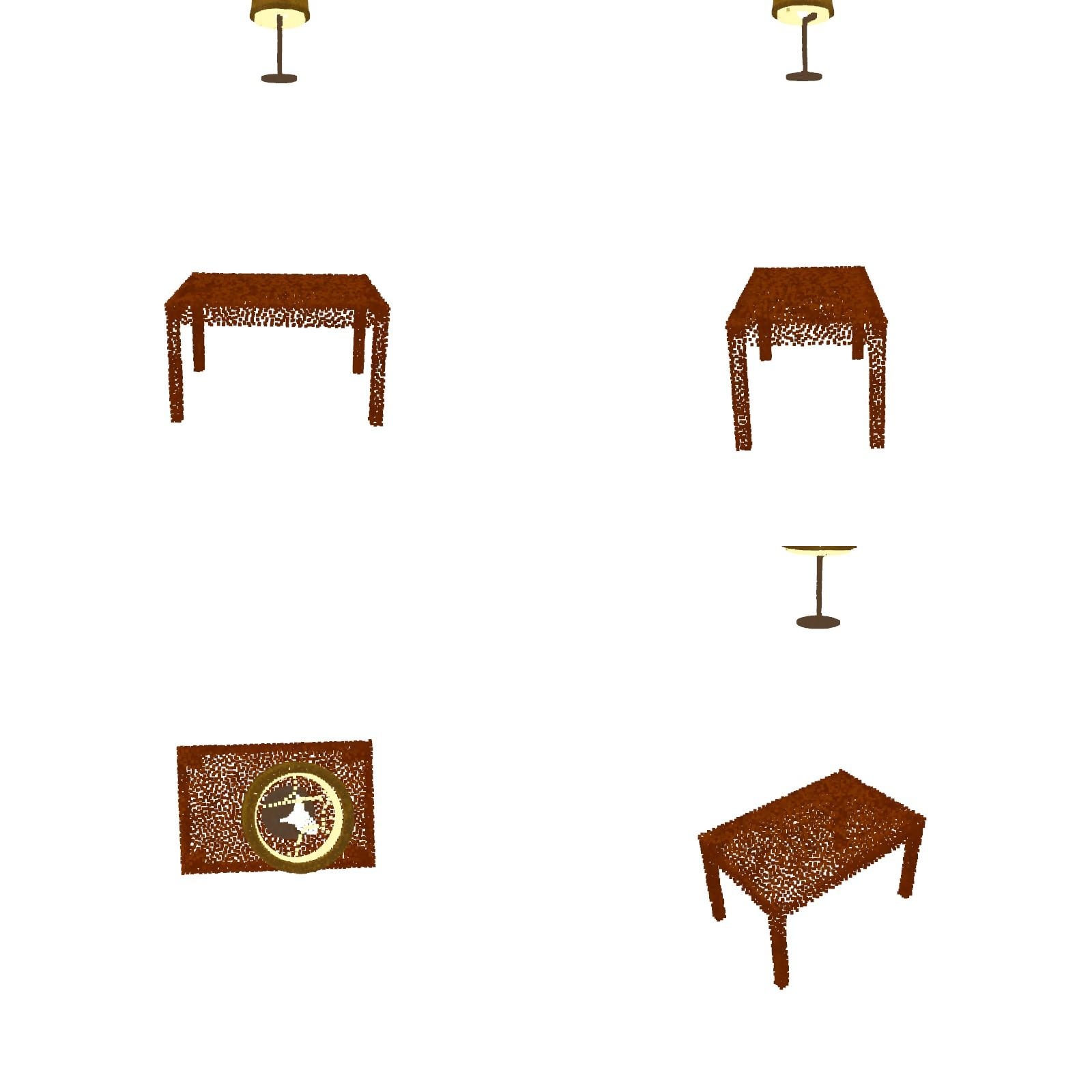}
   \caption{\textbf{Failure Case Example.} In rare cases, our method may encounter situations where objects move outside the camera’s capture range during the optimization process, leading to errors in subsequent computations. }
   \label{fig:appendix-failure}
 \end{figure}

\textbf{Independent Subgraph Optimization.} To preserve the integrity of relationships among existing nodes during the addition of new nodes, we adopt an Independent Subgraph Optimization approach. In this process, each newly optimized edge is incrementally integrated into the graph. If one of the nodes associated with the new edge already exists within the graph, its attributes are propagated to the other node. This propagation ensures consistency by facilitating numerical adjustments across connected nodes within the subgraph.

\textbf{LLM-Guided Subgraph Placement.} This method is designed to further refine the spatial positions of nodes within each subgraph. For every node in the graph, the corresponding subgraph is rendered, and its layout is optimized using an approach analogous to edge optimization but with tailored prompt information. During this process,  $X_1$  represents the specific node under optimization,  $X_2$  denotes the remaining nodes within the subgraph, and  $X_{\text{all}}$  encompasses all nodes within the subgraph. This ensures cohesive refinement of node positions while maintaining consistency within the subgraph structure.

\textbf{Graph Placement.} The Graph Placement phase focuses on optimizing subgraphs as complete units, expanding on their earlier individual optimizations. Like nodes, subgraphs are defined by attributes such as position, rotation, and scale. The process starts with an initial step where the LLMs estimate rough attribute values for each subgraph based on the descriptions of all subgraphs in the graph. These estimates serve as starting points for further optimization. Next, we refine these attributes through a process similar to edge optimization, where  $X_1$  is the subgraph being optimized,  $X_2$  includes all other subgraphs, and  $X_{\text{all}}$  represents the entire graph. This hierarchical optimization framework, progressing from edges to subgraphs and ultimately to the complete graph, ensures a cohesive and globally consistent structure.

\section{Failure Cases}

In our experiments, we identified occasional cases where our method encountered challenges during various stages of the optimization process. These issues were primarily due to ambiguities or inconsistencies in the input descriptions and rare errors in the scoring predictions by the MLLMs. Specifically, like the Figure ~\ref{fig:appendix-failure} during edge optimization, unclear edge descriptions or scoring inaccuracies sometimes resulted in inadequate multi-view capture of the objects associated with an edge, ultimately affecting the optimization outcome. This issue can be addressed by either imposing constraints to ensure object positions remain within the camera’s field of view or dynamically adjusting the camera positions to maintain consistent visibility of the objects throughout the optimization process.

\begin{table*}[ht]
\centering
\caption{Template Prompts Used in Our Method}
\label{tab:template_prompts}
\begin{tabular}{|p{1.5cm}|p{14.5cm}|} 
\hline
\textbf{Prompt Name} & \textbf{Prompt Content} \\ \hline
Prompt 1& You are an expert in computer graphics, computer vision, and scene design. Below I will send you a sentence. The sentence will describe some objects in a scene. I want you to help me construct a graph with nodes and edges, where nodes represent the objects in the scene, and edges represent the objects' connections. 

Here are the guided steps to construct the structure:

First, you should analyze the sentence, identify all object categories, count the objects, and assign each object a short prompt for 3D generation. These objects are the nodes of the graph. The result of nodes should follow this format: 
\texttt{"nodes = [obj\_1, obj\_2, obj\_3, \ldots], node-prompts = [prompt of obj\_1, prompt of obj\_2 , prompt of obj\_3]"} 
For example: 
\texttt{"nodes = [apple, banana, toy], node-prompts = [a fresh red apple, a ripe yellow banana, a colorful toy car]"}. 

Secondly, after collecting all nodes, you should identify all connections between objects. These connections are the edges of the graph, which should strictly be uni-directional. You should only use interaction like  \texttt{\{left, right, up, down, front, below, in\}} to describe the interactions between objects. The result of edges should follow this format, where \texttt{"obj\_a \{interaction\} obj\_b"} means \texttt{"obj\_a"} is in the position described by \texttt{"interaction"} relative to \texttt{"obj\_b"}: 
\texttt{"edges = [obj\_1 \{interaction\_1\} obj\_2, obj\_2 \{interaction\_4\} obj\_3, \ldots]"}. 
For example: 
\texttt{"edges = [apple left banana, toy on bed]"}. 

You should determine the most common interaction if there are multiple choices. 

The target sentence is: \textbf{$T_s$}\\ \hline
 Prompt 2&You are an expert in computer graphics, computer vision, and scene design. I will send you a sentence and3  images, all images are four views of a scene, where the left-top is a front view, the right-top is a side view, the bottom-left is a top-down view, and the bottom-right is an angled perspective view.

These images are respectively:

1. The optimized objects: $X_{1}$.
2. Other objects: $X_{2}$.
3. entire scene $X_{\text{all}}$.

The position, rotation, and scale of $X_{2}$ are correct in the scene. There might be some incorrect scale, location, and rotation of $X_{1}$, which are unlikely to form a realistic layout in a scene satisfying $X_{\text{all}}$ in the third image. Please now modify the scene step by step:

Please evaluate whether $X_{1}$ in the scene meets the requirement. Provide five scores from -100 to 100 based on the following criteria respectively:

1. First score is about the scale of $X_{1}$:
    If $X_{1}$ in the scene is at an appropriate size, give a score close to zero.
    If $X_{1}$ in the scene is too big, give a high positive score.
    If $X_{1}$ in the scene is too small, give a high negative score.
    
2. Second score is about $X_{1}$'s location in the left-and-right direction:
    (You must not consider the side view image to rate the score; consider the x-axis in both the front-view and top-down view.)
    If $X_{1}$ in the scene is at an appropriate location, give a score close to zero.
    If $X_{1}$ is too close to $X_{2}$, give a high positive score.
    If $X_{1}$ is too far from $X_{2}$, give a high negative score.

3. Third score is about $X_{1}$'s location in the forward-and-backward direction:
    (You must not consider the front-view image to rate the score; consider the x-axis in the side-view and the y-axis in the top-down view.)
    If $X_{1}$ in the scene is at an appropriate location, give a score close to zero.
    If $X_{1}$ is too close to $X_{2}$, give a high positive score.
    If $X_{1}$ is too far from $X_{2}$, give a high negative score.

4. Fourth score is about $X_{1}$'s location in the up-and-down direction:
    (You must not consider the top-down view to rate this score; consider the y-axis in both the front-view and side-view.)
    If $X_{1}$ in the scene is at an appropriate location, give a score close to zero.
    If $X_{1}$ is too close to $X_{2}$, give a high positive score.
    If $X_{1}$ is too far from $X_{2}$, give a high negative score.

5. Fifth score is about $X_{1}$'s yaw rotation:
    (You should consider the top-view image.)
    If $X_{1}$ in the scene is at an appropriate rotation, give a score close to zero.
    If $X_{1}$ should rotate clockwise, give a positive score.
    If $X_{1}$ should rotate counterclockwise, give a negative score.

The return should begin with:  
The score-1 is: ...  
The score-2 is: ...  
The score-3 is: ...  
The score-4 is: ...  
The score-5 is: ...\\\hline
\end{tabular}
\end{table*}

\end{document}